\documentclass[11pt,a4paper]{article}
\usepackage[hyperref]{acl2019}
\usepackage{times}
\usepackage{latexsym}
\usepackage{url}
\usepackage{graphicx}
\usepackage{amsmath}
\usepackage{amssymb}
\usepackage{subfigure}
\usepackage{caption}
\usepackage{algorithm}
\usepackage{algorithmicx}
\usepackage{algpseudocode}
\usepackage{multicol}
\usepackage{multirow}
\usepackage{booktabs}
\usepackage{listings}
\usepackage{array}

\aclfinalcopy 

\title{GEAR: Graph-based Evidence Aggregating and Reasoning \\ for Fact Verification}

\author{Jie Zhou$^{1,2,3}$, Xu Han$^{1,2,3}$, Cheng Yang$^{1,2,3}$, Zhiyuan Liu$^{1,2,3\dagger}$ \\ \textbf{Lifeng Wang$^{4}$, Changcheng Li$^{4}$, Maosong Sun$^{1,2,3}$}\\
\textsuperscript{1}Department of Computer Science and Technology, Tsinghua University, Beijing, China\\
\textsuperscript{2}Institute for Artificial Intelligence, Tsinghua University, Beijing, China\\
\textsuperscript{3}State Key Lab on Intelligent Technology and Systems, Tsinghua University, Beijing, China\\
\textsuperscript{4}Tencent Marketing Solution, Tencent, Shenzhen, China \\
\texttt{\{zhoujie18, hanxu17, cheng-ya14\}@mails.tsinghua.edu.cn} \\
\texttt{\{fandywang, harrychli\}@tencent.com, \{liuzy, sms\}@tsinghua.edu.cn}
}

\date{}

\begin{document}
\maketitle
\begin{abstract}
Fact verification (FV) is a challenging task which requires to retrieve relevant evidence from plain text and use the evidence to verify given claims. Many claims require to simultaneously integrate and reason over several pieces of evidence for verification. However, previous work employs simple models to extract information from evidence without letting evidence communicate with each other, e.g., merely concatenate the evidence for processing. Therefore, these methods are unable to grasp sufficient relational and logical information among the evidence. To alleviate this issue, we propose a graph-based evidence aggregating and reasoning (GEAR) framework which enables information to transfer on a fully-connected evidence graph and then utilizes different aggregators to collect multi-evidence information. We further employ BERT, an effective pre-trained language representation model, to improve the performance. Experimental results on a large-scale benchmark dataset FEVER have demonstrated that GEAR could leverage multi-evidence information for FV and thus achieves the promising result with a test FEVER score of 67.10\%. Our code is available at \url{https://github.com/thunlp/GEAR}.
\end{abstract}

\section{Introduction}
{\let\thefootnote\relax\footnotetext{$^\dagger$ Corresponding author: Z.Liu(liuzy@tsinghua.edu.cn)}}

Due to the rapid development of information extraction (IE), huge volumes of data have been extracted. How to automatically verify the data becomes a vital problem for various data-driven applications, e.g., knowledge graph completion~\cite{wang2017knowledge} and open domain question answering~\cite{chen2017reading}. Hence, many recent research efforts have been devoted to fact verification (FV), which aims to verify given claims with the evidence retrieved from plain text. More specifically, given a claim, an FV system is asked to label it as ``SUPPORTED'', ``REFUTED'', or ``NOT ENOUGH INFO'', which indicate that the evidence can support, refute, or is not sufficient for the claim.

\begin{table}[t]
\centering
\small
\scalebox{0.8}{
\begin{tabular}{l|p{0.9\columnwidth}}
\toprule
\multicolumn{2}{c}{``SUPPORTED'' Example}\\
\midrule
\multirow{1}{*}{Claim}        & The Rodney King riots took place in the most populous county in the USA.\\ \midrule
\multirow{2}{*}{Evidence}     & (1) The 1992 Los Angeles riots,  \textbf{\emph{\textcolor{red}{also known as the Rodney King riots}}} were a series of riots, lootings, arsons, and civil disturbances that  \textbf{\emph{\textcolor{red}{occurred in Los Angeles County}}}, California in April and May 1992. \\
                              & (2)   \textbf{\emph{\textcolor{red}{Los Angeles County}}}, officially the County of Los Angeles,  \textbf{\emph{\textcolor{red}{is the most populous county in the USA}}}. \\
\bottomrule
\toprule
\multicolumn{2}{c}{``REFUTED'' Example}\\
\midrule
\multirow{1}{*}{Claim}        & Giada at Home was only available on DVD.\\ \midrule
\multirow{2}{*}{Evidence}     & (1) \textbf{\emph{\textcolor{red}{Giada at Home}}} is a television show and first  \textbf{\emph{\textcolor{red}{aired}}} on October 18, 2008,  \textbf{\emph{\textcolor{red}{on the Food Network}}}. \\
                              & (2)  \textbf{\emph{\textcolor{red}{Food Network}}} is an American  \textbf{\emph{\textcolor{red}{basic cable and satellite television channel}}}. \\

\bottomrule
\end{tabular}
}
\caption{Some examples of reasoning over several pieces of evidence together for verification. The italic words are the key information to verify the claim. Both of the claims require to reason and aggregate multiple evidence sentences for verification.}
\label{tab:example}
\vspace{-0mm}
\end{table}

Existing FV methods formulate FV as a natural language inference (NLI)~\cite{angeli2014naturalli} task. 
However, they utilize simple evidence combination methods such as concatenating the evidence or just dealing with each evidence-claim pair.
These methods are unable to grasp sufficient relational and logical information among the evidence. In fact, many claims require to simultaneously integrate and reason over several pieces of evidence for verification. As shown in Table~\ref{tab:example}, for both of the ``SUPPORTED'' example and ``REFUTED'' example, we cannot verify the given claims via checking any evidence in isolation. The claims can be verified only by understanding and reasoning over the multiple evidence.

To integrate and reason over information from multiple pieces of evidence, we propose a graph-based evidence aggregating and reasoning (GEAR) framework.
Specifically, we first build a fully-connected evidence graph and encourage information propagation among the evidence. Then, we aggregate the pieces of evidence and adopt a classifier to decide whether the evidence can support, refute, or is not sufficient for the claim.
Intuitively, by sufficiently exchanging and reasoning over evidence information on the evidence graph, the proposed model can make the best of the information for verifying claims. For example, by delivering the information ``Los Angeles County is the most populous county in the USA'' to ``the Rodney King riots occurred in Los Angeles County'' through the evidence graph, the synthetic information can support ``The Rodney King riots took place in the most populous county in the USA''. Furthermore, we adopt an effective pre-trained language representation model BERT~\cite{devlin2018bert} to better grasp both evidence and claim semantics.

We conduct experiments on the large-scale benchmark dataset for Fact Extraction and VERification (FEVER)~\cite{thorne2018fever}. Experimental results show that the proposed framework outperforms recent state-of-the-art baseline systems. The further case study indicates that our framework could better leverage multi-evidence information and reason over the evidence for FV. 

\section{Related Work}
\subsection{FEVER Shared Task}
The FEVER shared task~\cite{thorne18fact} challenges participants to develop automatic fact verification systems to check the veracity of human-generated claims by extracting evidence from Wikipedia. The shared task is hosted as a competition on Codalab\footnote{\url{https://competitions.codalab.org/competitions/18814}} with a blind test set. \newcite{nie2018combining, yoneda2018ucl} and  \newcite{hanselowski2018ukp} have achieved the top three results among 23 teams.

Existing methods mainly formulate FV as an NLI task. \newcite{thorne2018fever} simply concatenate all evidence together, and then feed the concatenated evidence and the given claim into the NLI model. \newcite{luken2018qed} adopt the decomposable attention model (DAM)~\cite{parikh2016decomposable} to generate NLI predictions for each claim-evidence pair individually and then aggregate all NLI predictions for final verification. Then, \newcite{hanselowski2018ukp,yoneda2018ucl,hidey2018team} adopt the enhanced sequential inference model (ESIM)~\cite{chen2017enhanced}, a more effective NLI model, to infer the relevance between evidence and claims instead of DAM. As pre-trained language models have achieved great results on various NLP applications, \newcite{malon2018team} fine-tunes the generative pre-training transformer (GPT)~\cite{radford2018improving} for FV. Based on the methods mentioned above, \newcite{nie2018combining} specially design the neural semantic matching network (NSMN), which is a modification of ESIM and achieves the best results in the competition. Unlike these methods, \newcite{yin2018twowingos} propose the \textsc{TwoWingOS} system which trains the evidence identification and claim verification modules jointly. 

\subsection{Natural Language Inference}
The natural language inference (NLI) task requires a system to label the relationship between a pair of premise and hypothesis as entailment, contradiction or neutral. Several large-scale datasets have been proposed to promote the research in this direction, such as SNLI~\cite{bowman2015large} and Multi-NLI~\cite{williams2018broad}. These datasets have made it feasible to train complicated neural models which have achieved the state-of-the-art results~\cite{bowman2015large, parikh2016decomposable, sha2016reading, chen2017enhanced, chen2017recurrent, munkhdalai2017neural,  nie2017shortcut, conneau2017supervised, gong2017natural, tay2018compare, ghaeini2018dr}. It is intuitive to transfer NLI models into the claim verification stage of the FEVER task and several teams from the shared task have achieved promising results by this way.

\subsection{Pre-trained Language Models} 
Pre-trained language representation models such as ELMo~\cite{peters2018deep} and OpenAI GPT~\cite{radford2018improving} are proven to be effective on many NLP tasks. BERT~\cite{devlin2018bert} employs bidirectional transformer and well-designed pre-training tasks to fuse bidirectional context information and obtains the state-of-the-art results on the NLI task. In our experiments, we find the fine-tuned BERT model outperforms other NLI-based models on the claim verification subtask of FEVER. Hence, we use BERT as the sentence encoder in our framework to better encoding semantic information of evidence and claims.

\begin{figure*}[t]
    \centering
    \includegraphics[width=\linewidth]{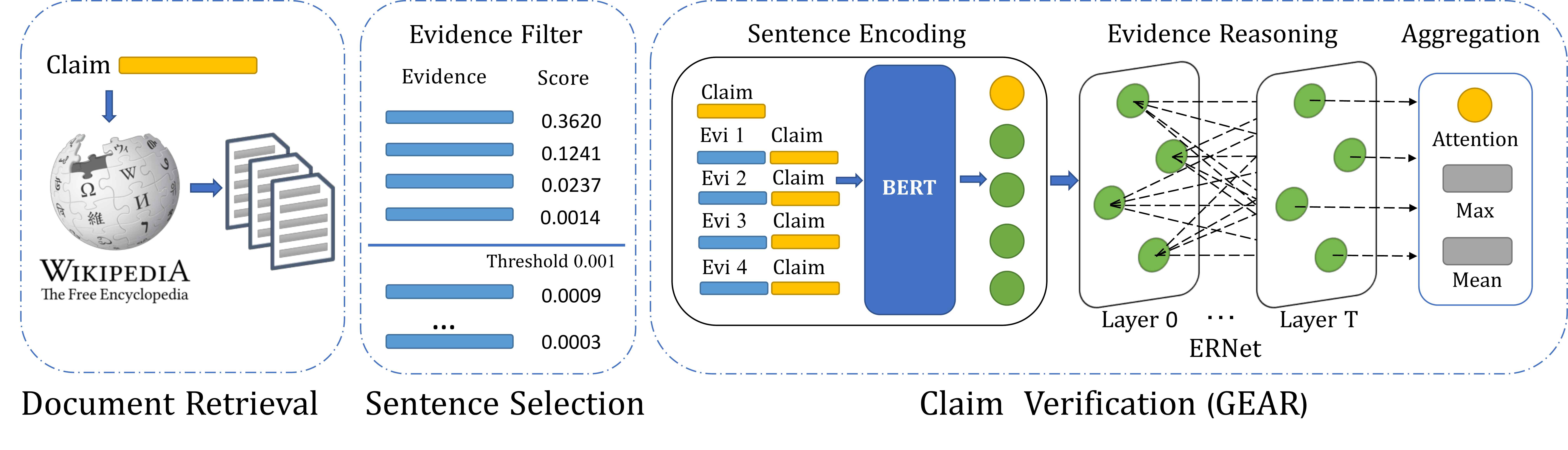}
    \caption{The pipeline of our method. The GEAR framework is illustrated in the claim verification section. }
    \label{fig:pipeline}
\end{figure*}

\section{Method}

We employ a three-step pipeline with components for document retrieval, sentence selection and claim verification to solve the task. In the document retrieval and sentence selection stages, we simply follow the method from~\newcite{hanselowski2018ukp} since their method has the highest score on evidence recall in the former FEVER shared task. And we propose our Graph-based Evidence Aggregating and Reasoning (GEAR) framework in the final claim verification stage. The full pipeline of our method is illustrated in Figure~\ref{fig:pipeline}.

\subsection{Document Retrieval and Sentence Selection}
In this section, we describe our document retrieval and sentence selection components. Additionally, we add a threshold filter after the sentence selection component to filter out those noisy evidence.

In the document retrieval step, we adopt the \emph{entity linking} approach from~\newcite{hanselowski2018ukp}. Given a claim, the method first utilizes the constituency parser from AllenNLP~\cite{gardner2018allennlp} to extract potential entities from the claim. Then it uses the entities as search queries and finds relevant Wikipedia documents via the online MediaWiki API\footnote{\url{https://www.mediawiki.org/wiki/API: Main_page}}. The seven highest-ranked results for each query are stored to form a candidate article set. Finally, the method drops the articles which are not in the offline Wikipedia dump and filters the articles by the word overlap between their titles and the claim. 

The sentence selection component selects the most relevant evidence for the claim from all sentences in the retrieved documents.

\newcite{hanselowski2018ukp} modify the ESIM model to compute the relevance score between the evidence and the claim. In the training phase, the model uses the hinge loss function $\sum max(0, 1 + s_n - s_p)$ with the negative sampling strategy, where $s_p$ and $s_n$ denote the relevance scores of positive and negative samples. In the test phase, the final model ensembles the results from 10 models with different random seeds. Sentences with top-5 relevance scores are selected to form the final evidence set in the original method.

In addition to the original model~\cite{hanselowski2018ukp}, we add a relevance score filter with a threshold $\tau$. Sentences with relevance scores lower than $\tau$ are filtered out to alleviate the noises. Thus the final size of the retrieved evidence set is equal to or less than 5.
We choose different values of $\tau$ and select the value based on the dev set result. The evaluation results of the document retrieval and sentence selection components are shown in Section~\ref{sec:dr-ss-evaluation}.

\subsection{Claim Verification with GEAR}

In this section, we describe our GEAR framework for claim verification. As shown in Figure~\ref{fig:pipeline}, given a claim and the retrieved evidence, we first utilize a \textbf{sentence encoder} to obtain representations for the claim and the evidence. Then we build a fully-connected evidence graph and propose an \textbf{evidence reasoning network} (ERNet) to propagate information among evidence and reason over the graph. Finally, we utilize an \textbf{evidence aggregator} to infer the final results.

\subsubsection*{Sentence Encoder} 
Given an input sentence, we employ BERT~\cite{devlin2018bert} as our sentence encoder by extracting the final hidden state of the [CLS] token as the representation, where [CLS] is the special classification embedding in BERT.

Specifically, given a claim $c$ and $N$ pieces of retrieved evidence $\{ e_1, e_2, ..., e_N \}$, we feed each evidence-claim pair $(e_i, c)$ into BERT to obtain the evidence representation $\mathbf{e_i}$. We also feed the claim into BERT alone to obtain the claim presentation $\mathbf{c}$. That is,
\begin{equation}
	\begin{split}
	\mathbf{e}_i &= \text{BERT}\left(e_i, c\right), \\	
	\mathbf{c} &= \text{BERT}\left(c\right).
	\end{split}
\end{equation}

Note that we concatenate the evidence and the claim to extract the evidence representation because the evidence nodes in the reasoning graph need the information from the claim to guide the message passing process among them.

\subsubsection*{Evidence Reasoning Network}
To encourage the information propagation among evidence, we build a fully-connected evidence graph where each node indicates a piece of evidence. We also add \emph{self-loop} to every node because each node needs the information from itself in the message propagation process. We use $\mathbf{h}^t = \{\mathbf{h}_1^t, \mathbf{h}_2^t, ..., \mathbf{h}_N^t\}$ to represent the hidden states of nodes at layer $t$, where $\mathbf{h}_i^t \in \mathcal{R}^{F \times 1}$ and $F$ is the number of features in each node. The initial hidden state of each evidence node $\mathbf{h}_i^0$ is initialized by the evidence presentation: $\mathbf{h}_i^0 = \mathbf{e}_i$.

Inspired by recent work on semi-supervised graph learning and relational reasoning~\cite{kipf2016semi,velickovic2017graph, palm2018recurrent}, we propose an evidence reasoning network (ERNet) to propagate information among the evidence nodes. We first use an MLP to compute the attention coefficients between a node $i$ and its neighbor $j$ ($j \in \mathcal{N}_i$),
\begin{equation}
	p_{ij} = \mathbf{W}_1^{t-1} ( \text{ReLU} ( \mathbf{W}_0^{t-1} ( \mathbf{h}_i^{t-1} \| \mathbf{h}_j^{t-1} ) ) ),
\end{equation}
where $\mathcal{N}_i$ denotes the set of neighbors of node $i$, $\mathbf{W}_0^{t-1} \in \mathcal{R}^{H \times 2F}$ and $\mathbf{W}_1^{t-1} \in \mathcal{R}^{1 \times H}$ are weight matrices, and $\cdot\|\cdot$ denotes concatenation operation.

Then, we normalize the coefficients using the softmax function,
\begin{equation}
	\alpha_{ij} = \text{softmax}_j(p_{ij}) = \frac{\text{exp}(p_{ij})}{\sum_{k \in \mathcal{N}_i} \text{exp}(p_{ik})}.
\end{equation}

Finally, the normalized attention coefficients are used to compute a linear combination of the neighbor features and thus we obtain the features for node $i$ at layer $t$,
\begin{equation}
	\mathbf{h}_i^t = \sum_{j \in \mathcal{N}_i} \alpha_{ij} \mathbf{h}_j^{t-1}.
\end{equation}

By stacking $T$ layers of ERNet, we assume that each evidence could grasp enough information by communicating with other evidence. We feed the final hidden states of evidence nodes $\{\mathbf{h}_1^T, \mathbf{h}_2^T, ...,\mathbf{h}_N^T\}$ into our evidence aggregator to make the final inference.

\subsubsection*{Evidence Aggregator}
We employ an evidence aggregator to gather information from different evidence nodes and obtain the final hidden state $\mathbf{o} \in \mathcal{R}^{F \times 1}$. The aggregator may utilize different aggregating strategies and we suggest three aggregators in our framework:

\textbf{Attention Aggregator.} Here we use the representation of the claim $\mathbf{c}$ to attend the hidden states of evidence and get the final aggregated state $\mathbf{o}$. 
	\begin{equation}
	\begin{split}
			p_{j} &= \mathbf{W}_1^{\prime} ( \text{ReLU} ( \mathbf{W}_0^{\prime} ( \mathbf{c} \| \mathbf{h}_j^{T} ) ) ), \\ 
			\alpha_{j} &= \text{softmax}(p_j) = \frac{\text{exp}(p_{j})}{\sum_{k=1}^N \text{exp}(p_{k})}, \\
			\mathbf{o} &= \sum_{k=1}^N \alpha_{k} \mathbf{h}_k^{T},
	\end{split}
	\end{equation}
where $\mathbf{W}_0^{\prime} \in \mathcal{R}^{H \times 2F}$ and $\mathbf{W}_1^{\prime} \in \mathcal{R}^{1 \times H}$.

\textbf{Max Aggregator.} The max aggregator performs the \emph{element-wise} Max operation among hidden states.
	\begin{equation}
		\mathbf{o} = \text{Max} (\mathbf{h}_1^T, \mathbf{h}_2^T, ..., \mathbf{h}_N^T).
	\end{equation}

\textbf{Mean Aggregator.} The mean aggregator performs the \emph{element-wise} Mean operation among hidden states.
	\begin{equation}
		\mathbf{o} = \text{Mean} (\mathbf{h}_1^T, \mathbf{h}_2^T, ..., \mathbf{h}_N^T).
	\end{equation}
	
Once the final state $\mathbf{o}$ is obtained, we employ a one-layer MLP to get the final prediction $l$.
\begin{equation}
	l = \text{softmax} (\text{ReLU}( \mathbf{W} \mathbf{o} + \mathbf{b})),
\end{equation}
where $\mathbf{W} \in \mathcal{R}^{C \times F}$ and $\mathbf{b} \in \mathcal{R}^{C \times 1}$ are parameters, and $C$ is the number of prediction labels.

\section{Experimental Settings}

\subsection{Dataset}
We conduct our experiments on the large-scale dataset FEVER~\cite{thorne2018fever}. The dataset consists of 185,455 annotated claims with a set of 5,416,537 Wikipedia documents from the June 2017 Wikipedia dump. We follow the dataset partition from the FEVER Shared Task~\cite{thorne18fact}. Table~\ref{tab:dataset} shows the statistics of the dataset.

\begin{table}[t!]
\begin{center}
\scalebox{0.85}{
\begin{tabular}{c  c  c  c}
\toprule \textbf{Split} & {SUPPORTED} & {REFUTED} & {NEI}\\ \midrule
Train & 80,035 & 29,775 & 35,639  \\
Dev & 6,666 & 6,666 & 6,666  \\
Test & 6,666 & 6,666 & 6,666  \\ \bottomrule
\end{tabular}}
\end{center}
\caption{\label{tab:dataset} Statistics of FEVER dataset.}
\end{table}

\subsection{Baselines}
In this section, we describe the baseline systems in our experiments. We first introduce the top-3 systems from the FEVER shared task.
As BERT~\cite{devlin2018bert} has achieved promising performance on several NLP tasks, we also implement two baseline systems via fine-tuning BERT in the claim verification task.

\subsubsection*{Shared Task Systems}
We choose the top-3 models from the FEVER shared task as our baselines. 

The Athene UKP TU Darmstadt team (Athene) \cite{hanselowski2018ukp} combines five inference vectors from the ESIM model via attention mechanism to make the final prediction. 

The UCL Machine Reading Group (UCL MRG)~\cite{yoneda2018ucl} predicts the label of each evidence-claim pair and aggregates the results via a label aggregation component. 

The UNC NLP team~\cite{nie2018combining} proposes the neural semantic matching network and uses the model jointly to solve all three subtasks. They also incorporate additional information such as pageview frequency and WordNet features. They have achieved best results in the competition.

\subsubsection*{BERT Fine-tuning Systems} 
We implement two BERT fine-tuning systems with different evidence combination approaches. The BERT-Concat system concatenates all evidence into a single string while the BERT-Pair system encodes each evidence-claim pair independently and then aggregates the results. Both systems share the same document retrieval and sentence selection components proposed by us.

\textbf{BERT-Concat.} In the BERT-Concat system, we simply concatenate all evidence into a single sentence and utilize BERT to predict the relation between the concatenated evidence and the claim. In the training phase, we add the ground truth evidence into the retrieved evidence set with relevance score 1 and select five pieces of evidence with the highest scores. In the test phase, we concatenate the retrieved evidence for predicting.

\textbf{BERT-Pair.} In the BERT-Pair system, we utilize BERT to predict the label for each evidence-claim pair. Concretely, we use each evidence-claim pair as the input and the label of the claim as the prediction target. 
In the training phase, we select the ground truth evidence for SUPPORTED and REFUTED claims and the retrieved evidence for NEI claims. In the test phase, we predict labels for all retrieved evidence-claim pairs.
Because different evidence-claim pairs may have inconsistent predicted labels, we then utilize an aggregator to obtain the final claim label. We find the aggregator only returning the predicted label from the most relevant evidence has the best performance. 

\subsection{Hyperparameter Settings}
We utilize $\text{BERT}_{\text{BASE}}$~\cite{devlin2018bert} in all of the BERT fine-tuning baselines and our GEAR framework. The learning rate is 2e-5.

For BERT-Concat, the maximum sequence length is 256 and the batch size is 16. We limit the max length for concatenated evidence to 240 and the max length for claims to 16. We train this model for two epochs based on dev results.
For BERT-Pair, we set the maximum sequence length to 128 and batch size to 32. We train this model for one epoch.
As for the GEAR framework, we use the fine-tuned BERT-Pair model to extract features and the batch size is 512.

In our ERNet, we set the batch size to 256, the number of features $F$ to 768 and the dimension of weight matrices $H$ to 64. The model is trained to minimize the negative log likelihood loss on the predicted label using the Adam optimizer~\cite{kingma2015adam} with an initial learning rate of 5e-3 and L2 weight decay of 5e-4. We use an early stopping strategy on the label accuracy of the validation set, with a patience of 20 epochs. We attempt to stack 0-3 ERNet layers and analyze the effect of different layer numbers.

\subsection{Evaluation Metrics}
Besides traditional evaluation metrics such as label accuracy and F1, we use other two metrics to evaluate our model.

\textbf{FEVER score.} The FEVER score is the label accuracy conditioned on providing at least one complete set of evidence. Claims labeled as ``NEI'' do not need the evidence. 

\textbf{OFEVER score.} The document retrieval and sentence selection components are usually evaluated by the oracle FEVER (OFEVER) score, which is the upper bound of the FEVER score by assuming perfect downstream systems.  
 
For all of the experiments with GEAR, the scores (label accuracy, FEVER score) we report on the dev set are mean values with $10$ runs initialized by different random seeds.

\begin{table}[t]
\begin{center}
\scalebox{0.80}{
\begin{tabular}{c  c}

\toprule \textbf{Model} & \textbf{OFEVER}  \\ \midrule
Athene & \textbf{93.55} \\ 
UCL MRG &  - \\ 
UNC NLP & 92.82  \\ 
Our Model & 93.33 \\ \bottomrule
\end{tabular}}
\end{center}
\caption{\label{tab:document-retrieval} Document retrieval evaluation on dev set (\%). ('-' denotes a missing value)}
\end{table}

\begin{table}[t]
\begin{center}
\scalebox{0.78}{
\begin{tabular}{c c c c c c}

\toprule \textbf{$\tau$} & \textbf{OFEVER} & \textbf{Precision} & \textbf{Recall} & \textbf{F1} & \textbf{GEAR LA} \\ \midrule
0 & \textbf{91.10} & 24.08 & \textbf{86.72} & 37.69 & 74.84\\
$10^{-4}$ & 91.04 & 30.88 & 86.63 & 45.53 & 74.86\\
$10^{-3}$ & 90.86 & 40.60 & 86.36 & 55.23 &  \textbf{74.91}\\
$10^{-2}$ & 90.27 & 53.12 & 85.47 & 65.52 & 74.89\\
$10^{-1}$ & 87.70 & \textbf{70.61} & 81.64 & \textbf{75.72} & 74.81\\ 
\bottomrule
\end{tabular}}
\end{center}
\caption{\label{tab:ss-cv-evaluation} Sentence selection evaluation and average label accuracy of GEAR with different thresholds on dev set (\%).}
\end{table}

\section{Experimental Results and Analysis}
In this section, we first present the evaluations of the document retrieval and sentence selection components. Then we evaluate our GEAR framework in several different aspects. Finally, we present a case study to demonstrate the effectiveness of our framework.

\subsection{Document Retrieval and Sentence Selection}
\label{sec:dr-ss-evaluation}
We use the OFEVER metric to evaluate the document retrieval component.
Table~\ref{tab:document-retrieval} shows the OFEVER scores of our model and models from other teams. After running the same model proposed by ~\newcite{hanselowski2018ukp}, we find our OFEVER score is slightly lower, which may due to the random factors.

Then we compare our sentence selection component with different thresholds, as shown in Table~\ref{tab:ss-cv-evaluation}. We find the model with threshold 0 achieves the highest recall and OFEVER score. When the threshold increases, the recall value and the OFEVER score drop gradually while the precision and F1 score increase. The results are consistent with our intuition. If we do not filter out evidence, more claims could be provided with the full evidence set. If we increase the value of the threshold, more pieces of noisy evidence are filtered out, which contributes to the increase of precision and F1.

\subsection{Claim Verification with GEAR}
In this section, we evaluate our GEAR framework in different aspects. We first compare the label accuracy scores between our framework and baseline systems. Then we explore the effect of different thresholds from the upstream sentence filter. We also conduct additional experiments to check the effect of sentence embedding.  As there are nearly 39\% of claims require reasoning over multiple pieces of evidence, we construct a difficult dev subset and check the effectiveness of our ERNet for evidence reasoning. Finally, we make an error analysis and provide the theoretical upper-bound label accuracy of our framework.

\subsubsection*{Model Evaluation}
We use the label accuracy metric to evaluate the effectiveness of different claim verification models.
The second column of Table~\ref{tab:pipeline} shows the label accuracy of different models on the dev set. We find the BERT fine-tuning models outperform all of the models from the shared task, which shows the strong capacity of BERT in representation learning and semantic understanding. The BERT-Concat model has a slight improvement over BERT-Pair, which is 0.37\%.

Our final model outperforms the best BERT-Concat baseline by 1.17\%. As our framework provides a better way for evidence aggregating and reasoning, the improvement demonstrates that our framework has a better ability to integrate features from different evidence by propagating, analyzing and aggregating the features. 

\begin{table}[t]
\begin{center}
\scalebox{0.85}{
\begin{tabular}{c c c c}
\toprule \multirow{2}{*}{\textbf{ERNet Layers}} & \multicolumn{3}{c}{\textbf{Aggregator}} \\ \cmidrule{2-4}
 & \textbf{Attention} & \textbf{Max} & \textbf{Mean} \\ \midrule
0 & 66.17 & 65.36 & 65.03 \\ 
1 & 67.13 & 66.63 & 66.76 \\ 
2 & \textbf{67.44} & \textbf{67.24} & \textbf{67.56} \\ 
3 & 66.53 & 66.72 & 66.89 \\  \bottomrule
\end{tabular}}
\end{center}
\caption{\label{tab:difficult} Label accuracy on the difficult dev set with different ERNet layers and evidence aggregators (\%).}
\end{table}

\begin{table}[t]
\begin{center}
\scalebox{0.85}{
\begin{tabular}{c c c c}
\toprule \multirow{2}{*}{\textbf{ERNet Layers}} & \multicolumn{3}{c}{\textbf{Aggregator}} \\ \cmidrule{2-4}
 & \textbf{Attention} & \textbf{Max} & \textbf{Mean} \\ \midrule
0 & 77.12 & 76.95 & 76.30 \\
1 & 77.74 & 77.66 & 77.62 \\
2 & \textbf{77.82} & \textbf{77.66} & \textbf{77.73} \\
3 & 77.70 & 77.55 & 77.60 \\  \bottomrule
\end{tabular}}
\end{center}
\caption{\label{tab:enhanced} Label accuracy on the evidence-enhanced dev set with different ERNet layers and evidence aggregators (\%).}
\end{table}

\begin{table}[t]
\begin{center}
\scalebox{0.85}{
\begin{tabular}{c  c  c  c  c}
\toprule \multirow{2}{*}{\textbf{Model}} & \multicolumn{2}{c}{\textbf{Dev}} &  \multicolumn{2}{c}{\textbf{Test}} \\ \cmidrule{2-5}
 & \textbf{LA} & \textbf{FEVER} & \textbf{LA} & \textbf{FEVER} \\ \midrule
Athene & 68.49 & 64.74 & 65.46 & 61.58 \\ 
UCL MRG &  69.66 & 65.41 & 67.62 & 62.52\\ 
UNC NLP &  69.72 & 66.49 & 68.21 & 64.21 \\ \midrule
BERT Pair & 73.30 & 68.90 & 69.75 & 65.18\\ 
BERT Concat & 73.67 & 68.89 & 71.01 & 65.64 \\
Our pipeline & \textbf{74.84} & \textbf{70.69} & \textbf{71.60} & \textbf{67.10} \\ \bottomrule
\end{tabular}}
\end{center}
\caption{\label{tab:pipeline} Evaluations of the full pipeline. The results of our pipeline are chosen from the model which has the highest dev FEVER score (\%).}
\end{table}

\subsubsection*{Effect of Sentence Thresholds}
The rightmost column of Table~\ref{tab:ss-cv-evaluation} shows the results of our GEAR frameworks with different sentence selection thresholds. We choose the model with threshold $\tau=10^{-3}$, which has the highest label accuracy, as our \textbf{final model}. When the threshold increases from 0 to $10^{-3}$, the label accuracy increases due to less noisy information. However, when the threshold increases from $10^{-3}$ to $10^{-1}$, the label accuracy decreases because informative evidence is filtered out, and the model can not obtain sufficient evidence to make the right inference.

\subsubsection*{Effect of Sentence Embedding}
The BERT model we used in the sentence encoding step is fine-tuned on the FEVER dataset for one epoch. We need to find out whether the fine-tuning process or simply incorporating the sentence embeddings from BERT makes the major contribution to the final result. We conduct an experiment using a BERT model without the fine-tuning process and we find the final dev label accuracy is close to the result from a random guess. Therefore, the fine-tuning process rather than sentence embeddings plays an important role in this task. We need the fine-tuning process to capture the semantic and logical relations between evidence and the claim. Sentence embeddings are more general and cannot perform well in this specific task. So that we cannot just use sentence embeddings from other methods (e.g., ELMo, CNN) to replace the sentence embeddings we used here.

\subsubsection*{Effectiveness of ERNet}
In our observation, more than half of the claims in the dev dataset only need one piece of evidence to make the right inference. To verify the effectiveness of our framework on reasoning over multiple pieces of evidence, we build a difficult dev subset via selecting samples from the original dev set. For SUPPORTED and REFUTED classes, claims which can be fully supported by only one piece of evidence are filtered out. All of the NEI claims are selected because the model needs all of the retrieved evidence to conclude that there is ``NOT ENOUGH INFO''. The difficult subset contains 7870 samples, which includes more than 39\% of the dev set. 

We test our final model on the difficult subset and present the results in Table~\ref{tab:difficult}. We find our models with ERNet perform better than models without ERNet and the minimal improvement between them is 1.27\%. We can also discover from the table that models with 2 ERNet layers achieve the best results, which indicates that claims from the difficult subset require multi-step evidence propagation.
This result demonstrates the ability of our framework to deal with claims which need multiple evidence.

\begin{table}[t]
\centering
\small
\scalebox{0.95}{
\begin{tabular}{p{\columnwidth}}
\toprule
\textbf{Claim:} \\ \textbf{\emph{\textcolor{red}{Al Jardine}}} is an \textbf{\emph{\textcolor{red}{American rhythm guitarist}}}.\\ \midrule
\textbf{Truth evidence:} \\ 
\{Al Jardine, 0\}, \{Al Jardine, 1\} \\ \midrule
\textbf{Retrieved evidence:} \\  
\{Al Jardine, 1\}, \{Al Jardine, 0\}, \{Al Jardine, 2\}, \{Al Jardine, 5\}, \{Jardine, 42\} \\ \midrule
\textbf{Evidence:} \\
\specialrule{0em}{1pt}{1pt}(1) \textbf{\emph{\textcolor{red}{He is best known as the band's rhythm guitarist}}}, and for occasionally singing lead vocals on singles such as ``Help Me, Rhonda'' (1965), ``Then I Kissed Her'' (1965) and ``Come Go with Me'' (1978).  \\
\specialrule{0em}{1.5pt}{1.5pt}(2) \textbf{\emph{\textcolor{red}{Alan Charles Jardine}}} (born September 3, 1942) is \textbf{\emph{\textcolor{red}{an American musician}}}, singer and songwriter who co-founded the Beach Boys.  \\ 
\specialrule{0em}{1.5pt}{1.5pt}(3) In 2010, Jardine released his debut solo studio album, A Postcard from California.  \\
\specialrule{0em}{1.5pt}{1.5pt}(4) In 1988, Jardine was inducted into the Rock and Roll Hall of Fame as a member of the Beach Boys.  \\
\specialrule{0em}{1.5pt}{1.5pt}(5) Ray Jardine American rock climber, lightweight backpacker, inventor, author and global adventurer. \\
\midrule
\textbf{Label:} SUPPORTED \\
\bottomrule
\end{tabular}
}
\caption{A case of the claim that requires integrating multiple evidence to verify. The representation for evidence ``\{\emph{DocName}, \emph{LineNum}\}'' means the evidence is extracted from the document ``\emph{DocName}'' and of which the line number is \emph{LineNum}.}
\label{tab:case-study}
\end{table}

\subsubsection*{Error Analysis}
In this section, we examine the effect of errors propagating from upstream components. We utilize an evidence-enhanced dev subset, which assumes all pieces of ground truth evidence are retrieved, to test the theoretical upper-bound score of our GEAR framework.

In our analysis, the main errors of our framework come from the upstream document retrieval and sentence selection components which can not extract sufficient evidence for inferring. For example, to verify the claim ``Giada at Home was only available on DVD'', we need the evidence ``Giada at Home is a television show and first aired on October 18, 2008, on the Food Network.'' and ``Food Network is an American basic cable and satellite television channel.''. However, the entity linking method used in our document retrieval component could not retrieve the ``Food Network'' document only from parsing the content of the claim. Thus the claim verification component can not make the right inference with insufficient evidence. 

To explore the effect of this issue, we test our models on an evidence-enhanced dev set, in which we add the ground truth evidence with relevance score 1 into the evidence set before the sentence threshold filter. It ensures that each claim in the evidence-enhanced set is provided with the ground truth evidence as well as the retrieved evidence. 

The experimental results are shown in Table~\ref{tab:enhanced}. We can find that all scores in the table increase by more than 1.4\% compared to the original dev set label accuracy in Table~\ref{tab:pipeline} because of the addition of the ground truth evidence. Because of the assumption of oracle upstream components, the results in Table~\ref{tab:enhanced} indicate the theoretical upper bound label accuracy of our framework.

The results show the challenges in the previous evidence retrieval task, which could not be solved by current models. \newcite{nie2018combining} propose a two-hop evidence enhancement method which improves 0.08\% on their final FEVER score. As the addition of the ground truth evidence leads to a more than 1.4\% increase in our experiment, it is worthwhile to design a better evidence retrieval pipeline, which remains to be our future research.

\subsection{Full Pipeline}
We present the evaluation of our full pipeline in this section. Note that there is a gap between the label accuracy and the final FEVER score due to the completeness of the evidence set. We find that a model which is good at predicting NEI instances tends to obtain higher FEVER score. 
So we choose our final model based on the dev FEVER score among all of our experiments. 
This model contains one layer of ERNet and uses the attention aggregator. The threshold of the sentence filter is $10^{-3}$.

Table~\ref{tab:pipeline} presents the evaluations of the full pipeline. We find the test FEVER score of BERT fine-tuning systems outperform other shared task models by nearly 1\%. Furthermore, our full pipeline outperforms the BERT-Concat baseline by 1.46\% and achieves significant improvements.

\begin{figure}[t]
    \centering
    \includegraphics[width=\linewidth]{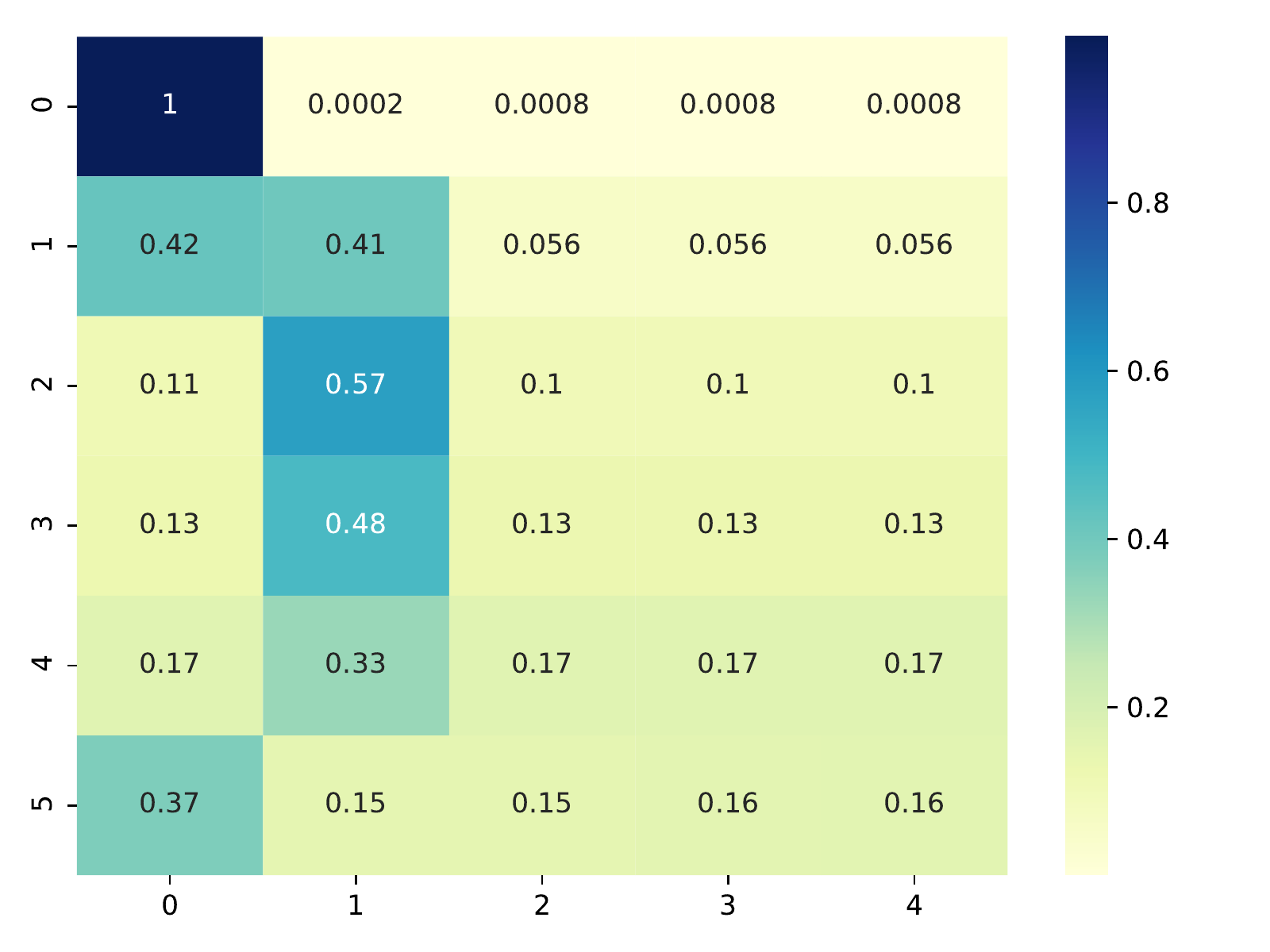}
    \caption{Attention map for the case in Table~\ref{tab:case-study}. The first five rows indicate the attention weights from nodes 1 to 5 in the first ERNet layer and the last row shows the attention weights from the attention aggregator.}
    \label{fig:case-attention}
\end{figure}

\subsection{Case study}

Table~\ref{tab:case-study} shows an example in our experiments which needs multiple pieces of evidence to make the right inference. 
The ground truth evidence set contains the sentences from the article ``Al Jardine'' with line number 0 and 1. These two pieces of evidence are also ranked at top two in our retrieved evidence set. To verify whether ``Al Jardine is an American rhythm guitarist'', our model needs the evidence ``He is best known as the band’s rhythm guitarist'' as well as the evidence ``Alan Charles Jardine ... is an American musician''.  We plot the attention map from our final model with one layer of ERNet and the attention aggregator in Figure~\ref{fig:case-attention}. We can find that all evidence nodes tend to attend the first and the second evidence nodes, which provide the most useful information in this case. The attention weights in other evidence nodes are pretty low, which indicates that our model has the ability to select useful information from multiple pieces of evidence.

\section{Conclusion}
We propose a novel Graph-based Evidence Aggregating and Reasoning (GEAR) framework on the claim verification subtask of FEVER. The framework utilizes the BERT sentence encoder, the evidence reasoning network (ERNet) and an evidence aggregator to encode, propagate and aggregate information from multiple pieces of evidence. The framework is proven to be effective and our final pipeline achieves significant improvements. In the future, we would like to design a multi-step evidence extractor and incorporate external knowledge into our framework.

\section{Acknowledgements}
This work is supported by the National Key Research and Development Program of China (No. 2018YFB1004503), the National Natural Science Foundation of China (NSFC No.61772302, 61572273). This work is also supported by 2018 Tencent Marketing Solution Rhino-Bird Focused Research Program No. 201808. Han is also supported by 2018 Tencent Rhino-Bird Elite Training Program.

\bibliography{acl2019}
\bibliographystyle{acl_natbib}

\end{document}